\documentclass[preprint,12pt]{elsarticle}
\usepackage{graphicx}
\usepackage{epstopdf}
\usepackage{amssymb}
\usepackage{amsmath}
\usepackage{amsthm}
\usepackage{amsfonts}
\usepackage{lineno}
\usepackage{hyperref}

\journal{Neurocomputing}

\begin{document}

\begin{frontmatter}

\title{An Efficient Two-Stage Sparse Representation Method}

\author[rvt]{Chengyu Peng}
\ead{roccypeng@gmail.com}
\author[rvt]{Hong Cheng \corref{cor1}}
\ead{hchenguestc@gmail.com}
\author[els]{Manchor Ko}
\ead{man961.great@gmail.com}

\cortext[cor1]{Corresponding author}
\address[rvt]{School of Automation Engineering, University of Electronic Science and Technology of China, Chengdu, 611731, China}
\address[els]{Activision Publishing, Inc., Santa Monica California, USA}

\begin{abstract}
There are a large number of methods for solving under-determined linear inverse problem. Many of them have very high time complexity for large datasets. We propose a new method called Two-Stage Sparse Representation (TSSR) to tackle this problem. We decompose the representing space of signals into two parts, the measurement dictionary and the sparsifying basis. The dictionary is designed to approximate a sub-Gaussian distribution to exploit its concentration property. We apply sparse coding to the signals on the dictionary in the first stage, and obtain the training and testing coefficients respectively. Then we design the basis to approach an identity matrix in the second stage, to acquire the Restricted Isometry Property (RIP) and universality property. The testing coefficients are encoded over the basis and the final representing coefficients are obtained. We verify that the projection of testing coefficients onto the basis is a good approximation of the signal onto the representing space. Since the projection is conducted on a much sparser space, the runtime is greatly reduced. For concrete realization, we provide an instance for the proposed TSSR. Experiments on four biometrics databases show that TSSR is effective and efficient, comparing with several classical methods for solving linear inverse problem.

\end{abstract}


\end{frontmatter}

\section{Introduction}
Linear inverse problems arise throughout engineering and the mathematical sciences. In most applications, these
problems are ill-conditioned or under-determined, so we must apply additional regularizing constraints in order to obtain interesting solutions. Most modern approaches use the sparsity of the solution as a regularizer
\cite{tropp2010computational, cheng2013sparse}. In this paper
we first give a brief view of these algorithms for sparse approximation. Then we propose a new two-stage sparse
representation method solving linear inverse problem.

Roughly there are four class of approaches to solve linear inverse problems: greedy, convex relaxation,
proximal and combinatorial methods.
Orthogonal Matching Pursuit (OMP) \cite{Pati1993OMP} is one of the important greedy algorithms. OMP finds one
atom at a time for approximating the solution of the $l_0$ problem:

\begin{equation}
\label{eqn_1}  \min_x\|x\|_0 \quad s.t. \quad \|y-\Phi x\|\leq\epsilon, \quad (P_0)
\end{equation}
where $y$ is a target signal, $\epsilon>0$ is some error tolerance. We refer to the vector $x$ as representing
coefficient of $y$ respect to the dictionary $\Phi$. We say $x$ is $K$-sparse when $\|x\|_0 \leq K$. The dictionary
$\Phi\in\mathbb{R}^{m\times N}$ is a real matrix whose columns have unit Euclidean norm: $\|\varphi_j\|_2$=1 for
$\ j =1,2,\ldots,N$.
OMP accumulates the vectors which have the least residual with the representing coefficients. The accuracy is restricted
since OMP does not consider that the multiple correlated atoms might be jointly selected.
The other greedy algorithms, including st-OMP \cite{Donoho2012STOMT}, ROMP \cite{Needell2007ROMP}, etc,
use $l_1$-norm to replace the NP-hard $l_0$-norm minimization:
\begin{equation}
\label{eqn_2}  \min_x\|x\|_1 \quad s.t. \quad \|y-\Phi x\|\leq\epsilon, \quad (P_1)
\end{equation}
They work well when $x$ is very sparse, but will deviate from the ideal solution of Eqn. (\ref{eqn_1})
when the number of non-zero entries in $x$ increases, as illustrated in the paper \cite{Plumbley2009Sparse}.

CoSaMP \cite{Needell2009cosamp}, as a  combinatorial method for solving $(P_1)$, is a widely used method which
avoids the pure greedy nature of OMP which can never remove any atom once they are selected.
It provides rigorous bounds on the runtime that are much better than the available results for interior-point methods \cite{Wright05Interior-point} - e.g. $L_1$-Magic \cite{C05l1-magic}.
The uniformity property of CoSaMP shows it can recover all signals given a fixed sampling matrix and the stability
property guarantees its success when the samples are contaminated with noise.

The convex relaxation methods, as another branch, relaxes the $(P_0)$ form by $(P_1)$. Basis Pursuit denoising (BPDN) \cite{Donoho2001BPDN}, solves a regularization problem with a trade-off between having a small residual and making coefficients simple in the $l_1$ sense.
Basis Pursuit (BP) series methods are far more complicated than OMP series, because these methods obtain the global solution of the optimal problem in each iteration.

$L_1$-Magic \cite{C05l1-magic} is a collection of matlab routines which are based on standard interior-point methods. One class of algorithms within reformulates linear inverse problem as the second-order cone program, and solve it with log-barrier method, which use conjugate gradient method as inner core.

Least-angle regression (LARS) \cite{Efron2004LARS}, as an active set method, performs model select to find the optimal point iteratively. It produces a full piecewise linear solution path, which is useful in cross-validation or similar attempts to tune the
model.

Among the proximal methods, Iterative Shrinkage-Thresholding Algorithm (ISTA) \cite{Daubechies2004ISTA} solves the variant of the problem $(P_1)$,
\begin{equation}
\label{eqn_3} \min 1/2||\Phi x-y||^2 + \lambda ||x||_1,
\end{equation}
where $\lambda$ is a regularization parameter. Roughly speaking, each iteration comprises of a multiplication
by $\Phi$ and its adjoint, along with a scalar shrinkage step on the obtained result. A short survey on the
applications of ISTA series can be found in \cite{Yin2008Bregman}. FISTA \cite{Beck2009FISTA}, as a quicker
version of ISTA, is proposed recently. Still FISTA needs many iterations for solving inverse problem if $\lambda$ is
small which is required for a good approximation of Eqn. (\ref{eqn_2}).

IHT \cite{Blumensath2007} is  based on the surrogate objective from \cite{Herrity2006}:
\begin{equation}
 C^S_{l_0} (y, \mathbf{z}) = \parallel\! x - \Phi y \!\parallel^2_2 + \lambda\parallel\! y \!\parallel_0 - \parallel\! \Phi y - \Phi\mathbf{z} \!\parallel^2_2
 +  \parallel\! y - \mathbf{z} \!\parallel^2_2
\end{equation}
For $\parallel\! \Phi \!\parallel_2 < 1$, the above is a majorisation of the $l_0$ regularized sparse coding objective function. Using a fixed threshold,
the author show IHT converges to a local minimum. It also needs many iterations and the per-iteration cost is about the same as Matching
Pursuit.

GISA \cite{Zuo2013} cleverly extended the soft-thresholding operator for $l_p$-norm
regularized sparse coding problem. Instead of a fixed threshold (e.g. $.5$ in IHT), the authors derived the following non-linear
equation for the threshold $\tau$:
\begin{equation}
 \tau^{GST}_P(\lambda) = (2\lambda (1-p))^{\frac{1}{2-p}}  + \lambda p(2\lambda (1-p) )^{\frac{p-1}{2-p}}
\end{equation}
using the above threshold GST can always find the correct solution to the $l_p$-minimization problem $\mathop{min}\limits_x \frac{1}{2}(y-x)^2 + \lambda|x|^p$.
The authors show one iteration of GISA is sufficient for image deconvolution. Hence it is very efficient.

To reduce the iterations for a proximal approach, the Linearized Bregman (LB) algorithm \cite{Yin2010Analysis} is produced which is
equivalent to gradient descent applied to a certain dual formulation. The analysis shows that LB has the exact
regularization property; namely, it converges to an exact solution of $(P_1)$ whenever its smooth
parameter $\alpha$ is greater than a certain value. The LB algorithm returns the solution to $(P_1)$
by solving the model:
\begin{equation}
\label{eqn_4} \min_x \|x\|_1+1/2\alpha \|x\|_2^2   \quad s.t. \quad   \Phi x = y,
\end{equation}

LB replaces the quadratic penalty in $(P_1)$ with a linear term and uses a mixture of $l_1$
and $l_2$ norm for the regularization. This key modification produces a strictly convex differentiable
objective function.

The LB method requires $O(\frac{1}{\epsilon})$ iterations to obtain an $\epsilon$-optimal solution, while Accelerated Linearized Bregman Method (ALB) \cite{Huang2011Accelerated} reduces the iteration complexity to $O(\frac{1}{\sqrt{\epsilon}})$ while requiring almost the same computational effort on each iteration.
ALB converges much quicker than LB on three types of sensing matrices generated by the $randn(m, n)$ function \cite{Huang2011Accelerated}, which are standard Gaussian matrix, Normalized Gaussian matrix and Bernoulli +1/-1 matrix. The other merit is that the relative errors obtained by ALB as a function of the iteration number are much smaller than LB does.

How to quickly represent data has been an open problem to deal with in academic and industrial area.

Many algorithms similar to above representative approaches adopt one-stage coding technique, which encodes the original samples in a large projected space. Among them many recast $(P_1)$ as a convex program with quadratic constraints, the computational cost for practical applications can be prohibitively high for large-scale problems. 
Honglak Lee \cite{Lee2007Efficient}, etc., however, proposed an efficient sparse coding algorithm, iteratively solving two convex optimization problems: an $L_1$ regularized least squares problem and an $L_2$-constrained least square problem.
In \cite{Liu2010Robust}, an online tracking algorithm with two stage optimization was proposed to jointly minimize the target reconstruction error and maximize the discriminative power by selecting a sparse set of features. It is very effective in handling a number of challenging sequences.
TSR \cite{He2010Two} proposed a robust and fast sparse representation method based on divide and conquer strategy. It divided the procedure of recognition into outlier detection stage and recognition stage.

FSR \cite{JiaBinHuang2010FSR}, first uses KSVD method \cite{Aharon05k-svd} to construct an dictionary for sparse representation, then applies OMP \cite{Pati1993OMP} to the dictionary to generate coefficients, and forms a reduced dictionary with the coefficients for the sparse coding. It runs much quicker than $L_1$-Magic solver, but the KSVD method assumes an overcomplete dictionary and has trivial problem when the dimension of the signal is larger than that of the dictionary.

As signals can be modelled by a small set of atoms in a dictionary, FSRM \cite{Peng2012Fast} exploits the property and shows that the $l_1$-norm minimization problem can be reduced from a large and
dense linear system to a small and sparse one. It exploits CoSaMP to generate sparse coefficients to do the next coding. Experimental results with image recognition indicate FSRM achieves double-digit gain in speed with comparable accuracy compared with the $L_1$-Magic solver, and solves the trivial problem FSR has.

The above two methods shares the same two-stage structure. The core of both is to design a projected sparsifying basis in the first stage to represent the signals, and do sparse coding in the second stage.
Motivated by it, we propose
a Two-Stage Sparse Representation (TSSR) method to design the basis for rapid speed. TSSR makes the sparse approximation computationally tractable without sacrificing stable convergence. It represents the data in a smaller space for discriminative representation and reduces the runtime dramatically.

The remainder of the paper is organized as follows. Section 2 presents the derivation of TSSR for verification and
the algorithm flowchart as well as the complexity analysis. Section 3 gives an example to facilitate the method and shows the experimental results with discussion, and finally Section 4 offers the conclusion.

\section{Two-Stage Sparse Representation}

We first describe the derivation of TSSR, which generates a new form of $(P_1)$, then give an example to show the performance of the method.

\begin{figure}[t]
\centering
 \includegraphics[width=0.9\textwidth]{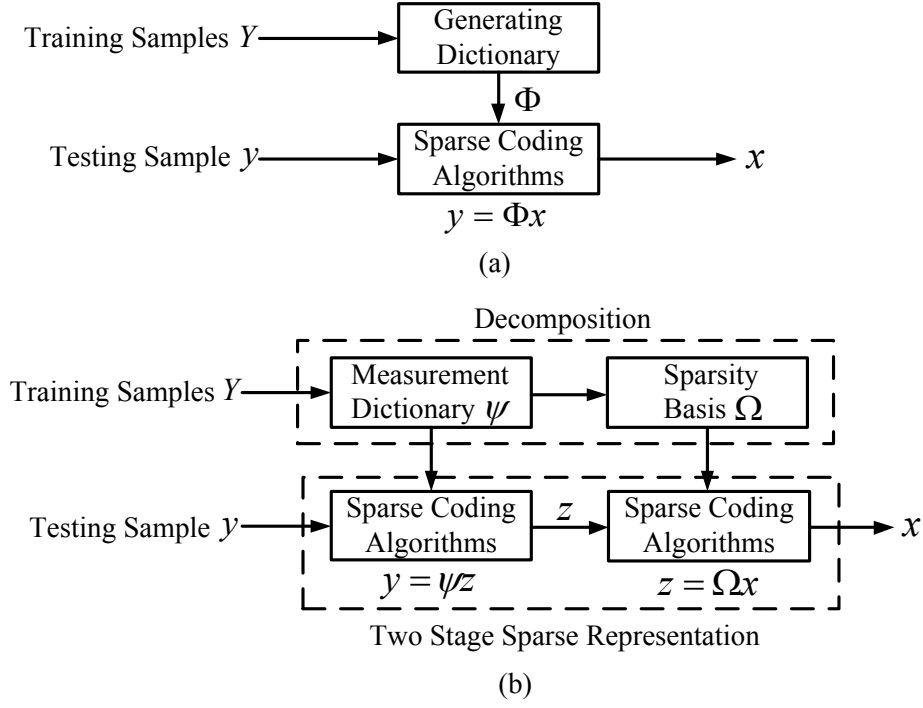}
\caption{The mechanism of one-stage sparse representation (a) and TSSR (b)}
\label{fig_1}
\end{figure}

\subsection{The Derivation of TSSR}
\label{Derivation}

The signal $y\in\mathbb{R}^{m}$ can be approximated in two ways:

1. $\Phi$ is viewed as a dictionary, $\Phi\in\mathbb{R}^{m\times N}$. $x \in\mathbb{R}^{N}$ is then the generated sparse coefficient vector over $\Phi$, and
\begin{equation}
\label{eqn_5} y=\Phi x+e_1,
\end{equation}
where $\Phi=[\phi_1,\ldots,\phi_i,\ldots,\phi_N]$.
$\Phi$ is supposed to satisfy the Restricted Isometry Property (RIP) of order $2K$ \cite{Candes2006stable}, and $e_1$ is the residual.
The illustration of Eqn. (\ref{eqn_5}) is shown as Figure \ref{fig_1}(a).

2. We can generate a new space $\Psi\Omega$, here $\Psi$ is viewed as a measurement dictionary,
and $\Omega=[\omega_1,\ldots,\omega_i,\ldots,\omega_N]$ represents a sparsity basis or dictionary in which $\omega_i$ expresses the $i$-th sparse vector.
Then we have another description of the signal $y$ over the $\Psi\Omega$ other than Eqn. (\ref{eqn_5}).
The mechanism of TSSR is illustrated in Figure \ref{fig_1}(b). To minimize the residual of the signal $\phi_i$, $\min\sum_{i=1}^{N}\|\phi_i-\Psi\omega_i\|^2$,
whose another form is $\min\|\Phi-\Psi\Omega\|^2$, we have
\begin{equation}
 \label{eqn_6} \Phi=\Psi\Omega +e_2.
\end{equation}
Here $\Psi\in\mathbb{R}^{m\times N}$, $\Omega\in\mathbb{R}^{N\times N}$, and $e_2$ is
supposed to obey the Gaussian distribution in the new space $\Psi\Omega$ for sparse reconstruction.
Then we can obtain
\begin{equation}
\label{eqn_7} y=\Psi\Omega x+e_2 x.
\end{equation}
After measuring the signal $y$, we can describe the intermediate representing coefficients $z$ as
\begin{equation}
\label{eqn_8}  z=\Omega x+e_3.
\end{equation}
$e_3$ is a small constant, while in Eqn. (\ref{eqn_6}) $\|e_2\|_2\leq\sqrt{N}\epsilon$ $(e_2\in\mathbb{R}^{m\times N})$.
By introducing the residual signals to  Eqn. (\ref{eqn_8}) and integrating Eqn. (\ref{eqn_7}), we can obtain
\begin{equation}
 \label{eqn_9}
 \begin{split} & \Psi z+e_4=\Psi\Omega x+e_2 x\\
 \Rightarrow & \Psi z-\Psi\Omega x=e_2 x-e_4.
 \end{split}
\end{equation}
Here $e_4$ is the same residual as $e_1$ in Eqn. (\ref{eqn_5}) and $\|e_4\|_2\leq\epsilon$, for a small constant $\epsilon$.
With \emph{bi-Lipschitz} property \cite{davenport2010signal}, we can derive
\begin{equation}
\label{eqn_10} (1-\delta)\|z-\Omega x\|_2^2\leq\|\Psi z-\Psi\Omega x\|_2^2\leq(1+\delta)\|z-\Omega x\|_2^2,
\end{equation}
given $\{\Omega x\}\in\mathbb{R}^N$. In fact, assume $x$ is $K$-sparse, from Eqn. (\ref{eqn_9}), we have
\begin{equation}
 \label{eqn_11} \|\Psi(z-\Omega x)\|_2\leq(K+1)\epsilon.
\end{equation}
Let $s=z-\Omega x$, which is also a sparse vector, we have $\|\Psi s\|_2\leq(K+1)\epsilon$. Recall that $\Psi$
satisfies the RIP of order $K$ with constant $\delta_K<1$ if
\begin{equation}
 \label{eqn_12} \|s\|_0\leq K\Rightarrow(1-\delta_K)\|s\|_2^2\leq\|\Psi s\|_2^2\leq(1+\delta_K)\|s\|_2^2.
\end{equation}
We can derive the upper bound for $s$ using Eqn. (\ref{eqn_11}) and Eqn. (\ref{eqn_12}).
\begin{equation}
\label{eqn_13} (1-\delta_K)\|s\|_2^2\leq\|\Psi s\|_2^2\leq(K+1)^2\epsilon^2.
\end{equation}
From Eqn. (\ref{eqn_13}) we have
\begin{equation}
 \label{eqn_14} \|s\|_2=\|z-\Omega x\|_2\leq\frac{(K+1)\epsilon}{\sqrt{1-\delta_K}}=\tilde{\epsilon}.
\end{equation}
Then the projection of $z$ onto $\Omega$ as shown in Figure \ref{fig_1}(b) is the approximation to that of $y$ onto $\Psi$.

So the original problem $(P_1)$ is modified by
\begin{equation}
 \label{eqn_15} \min_x\|x\|_1 \quad  s.t. \quad \|\Psi(z-\Omega x)\|\leq\epsilon_{1}
\end{equation}

Since the $\|\Psi\|$ is a constant, which has little effect on the solution of the object function, Eqn. (\ref{eqn_15})
then can be approximated by the following equation

\begin{equation}
 \label{eqn_16} \min_x\|x\|_1 \quad  s.t. \quad \|z-\Omega x\|\leq\epsilon_{2}
\end{equation}
$\epsilon_{1}$ and $\epsilon_{2}$ are small different constants.

Since the projection at the second stage is conducted on a much sparser space $\Omega$ than $\Phi$ in the first one, the runtime is greatly reduced.
Also the derivation above verifies the stability of TSSR, which satisfies the RIP condition under certain assumption.

The space $\Phi$ is approximated by the space $\Psi\Omega$ as Eqn. (\ref{eqn_6}) indicates, and the $\Psi$ and $\Omega$ are designed as follows:

We can design $\Psi$ to obey or nearly obey Gaussian distribution with the training data through matrix computation. Since Gaussian distribution with bounded support is Sub-Gaussian \cite{davenport2010signal}, we can exploit the concentration property which only requires Sub-Gaussian.

Any distribution satisfying a concentration inequality \cite{Talagrand1996Newlook} will provide both the canonical RIP and the universality with respect to a certain sparsity basis $\Omega$. Here $\Omega$ is represented as an identity $N\times N$ matrix and $x\in\Sigma_K$, each $K$-dimensional subspace from $\sum_K=\sum_K(\Omega)$ is mapped to a unique $K$-dimensional hyperplane in $\mathbb{R}^N$.
Once $\sum_K$ has a sufficient amount of independence, the concentration of measure tends to be sub-gaussian in nature. Then we can acquire signals $\Omega x$ that are sparse or compressible in practice. So $\Omega$ is designed to approach an identity matrix after the first-stage implementation to have the property described above.
By choosing $K$-dimensional subspaces spanned by sets of $K$ columns of $\Psi$, theorem 5.2 of \cite{baraniuk2008simple} establishes the RIP for $\Psi\Omega$ for each of the distributions.

The transformation from the first stage to the second one can be motivated by Concentration of Measure
principle \cite{Talagrand1996Newlook} and bi-Lipschitz theory \cite{davenport2010signal}.
Further derivations are found in the Appendix.

\subsection{The Algorithm Flowchart of TSSR }

The flowchart of TSSR is described as follows,

\begin{enumerate}[1.]
\item  Input: A test signal $y\in\mathbb{R}^{m}$, training signals $Y\in\mathbb{R}^{m\times N}$,
sparsity level $K$, a measurement matrix
$\Psi\in\mathbb{R}^{m\times N}$ with each column normalized.

\item  Generate the features of $Y$ over $\Psi$, using Eqn. (\ref{eqn_6}), and form $\Omega$ to approach an identity matrix, $\Omega=[\omega_1,\ldots,\omega_i,\ldots,\omega_N] \in\mathbb{R}^{N\times N}$, in which $\omega_i$ expresses the $i$-th sparse vector.
\item  Obtain the feature $z$ of $y$ over $\Psi$, using $y=\Psi z+e_4$,
where $z\in\mathbb{R}^{N}$.

\item  Reformulate the objective function $(P_1)$ as Eqn. (\ref{eqn_16}).
\item  Apply sparse coding to obtain representing coefficients $x\in\mathbb{R}^{N}$ over $\Omega$.
\end{enumerate}

Since the $\Omega$ is contained in a much sparser space than $\Psi\Omega$ does, the implementing speed increases dramatically, while the solution $x$ approaches the original one solving $(P_1)$.

\subsection{Complexity Analysis}

We give a brief description of the complexity analysis comparison between solving Eqn. (\ref{eqn_2}) and Eqn. (\ref{eqn_16}). The computational cost for the former hinges on $\Phi$ representing $y$, while that for the latter lies in $\Omega$ representing $z$.
For the $\Phi$ is contained in the space generated by the original data, the implementation over it completely depends on the ability of coding. Differently, the $\Omega$ can be designed to approach an identity matrix or a sparse matrix which has maximum value at the diagonal and small number of minor values off-diagonal of it. To do the coding task with the same optimization method, the time to take one sweep over the columns of $\Phi$ and $\Omega$ then need quite different cost. For instance, if we use CoSaMP \cite{Needell2009cosamp} for coding, we need $O(m\times N)$ and nearly $O(N)$ flops
for $\Phi$ and $\Omega$ respectively.
As a result, the recognition rate of the proposed TSSR is much quicker than many one-stage methods.

To demonstrate the effectiveness of the method, we give an application to the method in the next section.

\section{Experimental Results and Discussion}
\subsection{An Instance of TSSR }

An instance of TSSR is shown in the following. In the first stage, we adopt the ALB \cite{Huang2011Accelerated} to generate sparse coefficients of samples over $\Psi$, which has merits of very small relative errors, and achieves the desired convergence with small number of iterations. $\Psi$ can be formed by the training dataset $Y$ which are normalized, as shown in Figure \ref{fig_1}, and we can make $\Psi$ obey or nearly obey the Gaussian distribution.
The sparse coefficients (features) $\Omega$ corresponds to training data $Y$, and forms a square matrix (dictionary), in which each column is composed of the representatives of one sample. With sparse coding by ALB for different datasets, the generated $\Omega$ is similar to an identity matrix or a sparse matrix which has maximum value at the diagonal and relatively small values off-diagonal of it.

Then we use CoSaMP \cite{Needell2009cosamp} to acquire the sparse coefficients $z$ of test data $y$ over $\Psi$ and then represent $z$ over $\Omega$.
The uniformity property of CoSaMP shows it can recover all signals given a fixed sampling matrix and the stability property guarantees its success in solving problem $(P_1)$ when the samples are contaminated with noise. CoSaMP performs signal estimation and residual update, and then generates $K$ (sparsity level) non-zero coefficients for $z$ \cite{Candes2006stable}.
As section \ref{Derivation} shows, the problem $(P_1)$ then becomes searching for the sparsest solution on the
basis $\Omega$ as Eqn. (\ref{eqn_16}) shows.

For classification we use the sparse representation classifier (SRC) \cite{Wright2009PAMI} which is known to
have good robustness against signal corruption and noise. SRC minimizes the residuals between the test sample and
training samples of different classes, and find the label of test sample which corresponds to the minimum residual.

Note that TSSR structure can accommodate other sparse solvers as well as different measure matrices.

We present experimental results on real data sets to demonstrate the efficiency and effectiveness of the proposed algorithm. All the experiments were carried out using MATLAB on a 3.0GHz machine with 2G RAM. The time to classify one image is averaged over 10 runs. The bold values indicate
the best performances under specific condition. The parameter $\alpha$ adopted in ALB depends on the data \cite{Lai2013Augmented},
but a typical value is 1 to 10 times the estimate of  $\|x\_true\|_\infty$. Here we assume that an observed sample
belongs to one certain class and can be well represented using samples from the same class.

We present image recognition results with our TSSR in comparison with several benchmark methods solving linear inverse problem: matching pursuit method (CoSaMP), interior point method ($L_1$-Magic), active set method (Homotopy \cite{Donoho08fastsolution}), proximal method (FISTA), bregman method (ALB), and two-stage structure methods (FSR and FSRM).

\subsection{Face Recognition}
\subsubsection{PIE Database}

The CMU PIE database \url{http ://vasc.ri.cmu.edu/idb/html/face/} contains 68 human subjects with 41,368 face images as a whole. We choose the five near frontal poses (C05, C07, C09, C27, and C29) and use all the images under different illuminations and expressions, thus we get 170 images for each individual. 
Each image is manually aligned to $32 \times 32$ according to the eyes positions, with 256 gray levels per pixel. So each image can be represented by a 1024-dimensional vector in image space. No further preprocessing is done.
For PIE, we randomly select 1, 3, 6 samples for training, and the other 30 for test (i.e., cases 1-30, 3-30, 6-30) to evaluate the performance. The results are shown in Table 1.

\begin{table}
\caption{Recognition accuracy and speed comparison using PIE database.}
\begin{center}
\begin{tabular}{|c|c|c|c|c|c|c|}
\hline { \mbox{Samples}} & \multicolumn{2}{|c|}{PIE (1-30)} & \multicolumn{2}{|c|}{PIE (3-30)} & \multicolumn{2}{|c|}{PIE (6-30)} \\
\hline
Method & Acc. (\%) & T. (s) & Acc. (\%) & T. (s) & Acc. (\%) & T. (s)  \\
\hline \hline
TSSR & 73.09 & \textbf{4E-04} & 90.69 & \textbf{6E-04} & 92.25 & \textbf{0.001}  \\
Homotopy & 66.23 & 0.002 & 86.37 & 0.005 & 91.27 &  0.008  \\
CoSaMP & 72.65 & 0.004 & 89.56 & 0.007 & 93.09 & 0.009  \\
$L_1$-Magic & \textbf{75.69} & 0.005 & \textbf{94.36} & 0.019 & \textbf{95.49} & 0.14  \\
ALB & 74.85 & 0.091 & 92.6 & 0.148 & 93.58 & 0.24 \\
FISTA & 75.1 & 0.093 & 92.45 & 0.198 & 94.8 & 0.424  \\
\hline
\end{tabular} \\
\end{center}
\end{table}

As you see in table 1, TSSR is the fastest among all the methods. The rest of the methods except Homotopy need ten
times longer than TSSR does for all the cases, and FISTA even needs 230 times more to run. TSSR is also more
accurate than Homotopy for all the cases. The reason that Homotopy runs quickly is due to the fact that it iteratively
adds or removes nonzero representing coefficients one at a time, and is clearly more efficient when the signal
is very sparse. The time ratio of $L_1$-Magic, achieving the highest accuracy for all cases, to TSSR, changes rapidly
as the training samples becomes larger, from 12.5 to 140. This may due to the fact that least square method
within it needs much more time to run when the number of samples changes.

\subsubsection{AR Database}
The AR database \cite{Martinez2001AR} we choose contains 2600 color images corresponding to 100 people's faces (50 men and 50 women). Images feature frontal view faces with different facial expressions, illumination conditions, and occlusions (sun glasses and scarf). Each person participated in two sessions, separated by two weeks time.
 The size of the each image is 120x165 pixels, and was reduced to $40 \times 55$, i.e. a 2200 dimension vector.
Similarly, for AR, we randomly select 1, 3, 7 samples for training, and the rest for test , we use cases 1-13, 3-11, 7-7. The results are shown in Table 2.

\begin{table}
\caption{Recognition accuracy and speed comparison using AR database.}
\begin{center}
\begin{tabular}{|c|c|c|c|c|c|c|}
\hline { \mbox{Samples}} & \multicolumn{2}{|c|}{AR (1-13)} & \multicolumn{2}{|c|}{AR (3-11)} & \multicolumn{2}{|c|}{AR (7-7)} \\
\hline
Method & Acc. (\%) & T. (s) & Acc. (\%) & T. (s) & Acc. (\%) & T. (s)  \\
\hline \hline
TSSR & 70.31 & \textbf{8E-04} & 80.64 & \textbf{0.001} & 97 & \textbf{0.005}  \\
Homotopy & 73.85 & 0.044 & 88.36 & 0.032 & 96 &  0.0052  \\
CoSaMP & 71.92 & 0.023 & 89.55 & 0.041 & 97.29 & 0.083  \\
$L_1$-Magic & 69.31 & 0.011 & 90.91 & 0.304 & 98.71 & 0.068  \\
ALB & 71.15 & 0.583 & 86.91 & 1.577 & 99.14 & 3.231 \\
FISTA & \textbf{75.46} & 0.565 & \textbf{93} & 2.333 & \textbf{99.57} & 5.825  \\
\hline
\end{tabular} \\
\end{center}
\end{table}

TSSR is the fastest among all the methods as shown in table 2. For all the cases, ALB and FISTA can both obtain quite high accuracy but at high cost. For example, ALB (1.5771s) needs 1120 times more to run than TSSR (0.0014s) does for case (3-11). Both ALB and FISTA have high accuracy, this maybe because ALB has a small relative error, while FISTA is a proximal method with good convergence property. On the other side of view, the performance of TSSR on AR database is overall worse than the others, this may because
the sparse basis $\Omega$ constructed can not expressed the space information of the images very well.

\subsection{ Palmprint Recognition}

The PolyU palmprint database \cite{Zhang2003Online} contains 386 palms and each palm has about 20 samples, collected in two sessions
 separated by two months. The size of the each image is $384 \times 284$ pixels, and was reduced to dimension 1131. All the data are normalized.
For PolyU palmprint, we uses cases 1-10, 3-10, and 6-10. The results are shown in Table 3.

\begin{table}
\caption{Recognition accuracy and speed comparison using PolyU database.}
\begin{center}
\begin{tabular}{|c|c|c|c|c|c|c|}
\hline { \mbox{Samples}} & \multicolumn{2}{|c|}{PolyU (1-10)} & \multicolumn{2}{|c|}{PolyU (3-10)} & \multicolumn{2}{|c|}{PolyU (6-10)} \\
\hline
Method & Acc. (\%) & T. (s) & Acc. (\%) & T. (s) & Acc. (\%) & T. (s)  \\
\hline \hline
TSSR & 85.4 & \textbf{0.002} & 98.6 & \textbf{6E-04} & 99 & \textbf{0.001}  \\
Homotopy & 84 & 0.002 & 97 & 0.002 & 98.8 &  0.005  \\
CoSaMP & 86.6 & 0.005 & 98.4 & 0.004 & \textbf{99.2} & 0.006  \\
$L_1$-Magic & \textbf{86.8} & 0.005 & 98.2 & 0.027 & 98.4 & 0.197  \\
ALB & 85.8 & 0.074 & \textbf{98.8} & 0.12 & 99 & 0.205 \\
FISTA & 84.8 & 0.073 & 98 & 0.169 & 99 & 0.317  \\
\hline
\end{tabular} \\
\end{center}
\end{table}

 We can see in table 3 TSSR is superior to other methods in speed. Homotopy
 and CoSaMP have comparable speed, and run faster than the rest of the methods except TSSR. For CoSaMP,
 the RIP ensures that the least-square problems encountered are always well conditioned, so the iteration
 converges quickly. Interestingly the accuracy of TSSR is nearly the same as the methods with the highest accuracy in all the cases.

From all the experiments with the three databases, the results indicate TSSR is effective and efficient.
We note that sparsity level $K$ should be carefully chosen when TSSR is implemented. In the following section, we'll consider how to pick a good $K$ for a dataset.

\subsection{Choosing Sparsity Level}

Before indicating how we choose $K$, we first describe the relation between sparsity and mutual coherence \cite{Strohmer2003Grassmannian}.
For a dictionary $\Psi$, $\Psi\in\mathbb{R}^{m\times N}$, its coherence is bounded by

\begin{equation}
\label{eqn_17} M_{min}=\sqrt{\frac{N-m}{m(N-1)}}=O(\frac{1}{\sqrt{m}}),
\end{equation}
Since every entry in the off-diagonal of $\Psi$ is at most $M$, this leads to the condition $(K-1)M<1$, then
\begin{equation}
\label{eqn_18} K\leq\frac{1}{M}+1=\sqrt{\frac{m(N-1)}{N-m}}+1.
\end{equation}
$M$ is the Mutual Coherence, which is a measure of how similar the columns in $\Psi$ are to each other. If $M$ is large (close to 1, since $0<M\leq1$), it implies atoms are highly correlated and will led to poor performance for sparse representation. Geometrically this implies we want the atoms to be as orthogonal to each other as feasible - i.e. to form a Grassmanian frame in the sense of Benedetto \cite{Benedetto2006Geometric}.

The Extended Yale B database \cite{Georghiades2001YaleB} consists of 2432 grey images of 38 subjects under 9 poses and 64 illumination conditions. We choose the frontal pose and use all the images under different illumination, thus we get 64 images for each person.
Each image is manually aligned to $32 \times 32$ according to the eyes positions, with 256 gray levels per pixel. So each image can be represented by a 1024-dimensional vector in image space. No further preprocessing is done.

For the $(132\times1209)$ dictionary $\Psi$, which we use in all the following experiments, the sparsity level is $K\leq13.2$ as prescribed by Eqn. (\ref{eqn_18}). To verify this method of estimating $K$, we begin our experiments with $K$ set to 5, 10, 13, 18, and 20 and study the various methods' performance.

\label{table4}
\begin{table}
\caption{The Accuracy and speed comparison of FSR, FSRM and TSSR using Extended Yale B database}
\begin{center}
\begin{tabular}{|c|c|c|c|c|c|c|c|}
\hline { \mbox{Item}} & \multicolumn{3}{|c|}{Acc. (\%)} & \multicolumn{3}{|c|}{T. (s)} & { \mbox{Speed}} \\
\cline{1-7}
Method & FSR &FSRM & TSSR & FSR & FSRM & TSSR  & {up} \\
\hline \hline
K1=5 & 83.15 & 90.87 & \textbf{94.19} & \textbf{0.011} & 0.049 & 0.014 & 1 \\
K2=10 & 90.79 & 91.2 & \textbf{93.20}& 0.163 & 0.052 &  \textbf{0.015} & 11 \\
K3=13 & 93.36 & 93.03 & \textbf{94.36} & 0.432 & 0.052 &  \textbf{0.014} & 31 \\
K4=18 & 92.61 & 90.62 & \textbf{93.53} & 0.755 & 0.054 &  \textbf{0.015} & 52 \\
K5=20 & 91.95 & 90.71 & \textbf{92.78} & 0.837 & 0.053 &  \textbf{0.014} & 61 \\
\hline
\end{tabular} \\
\end{center}
\end{table}

 \begin{figure}[t]
\centering
 \includegraphics[width=0.75\textwidth]{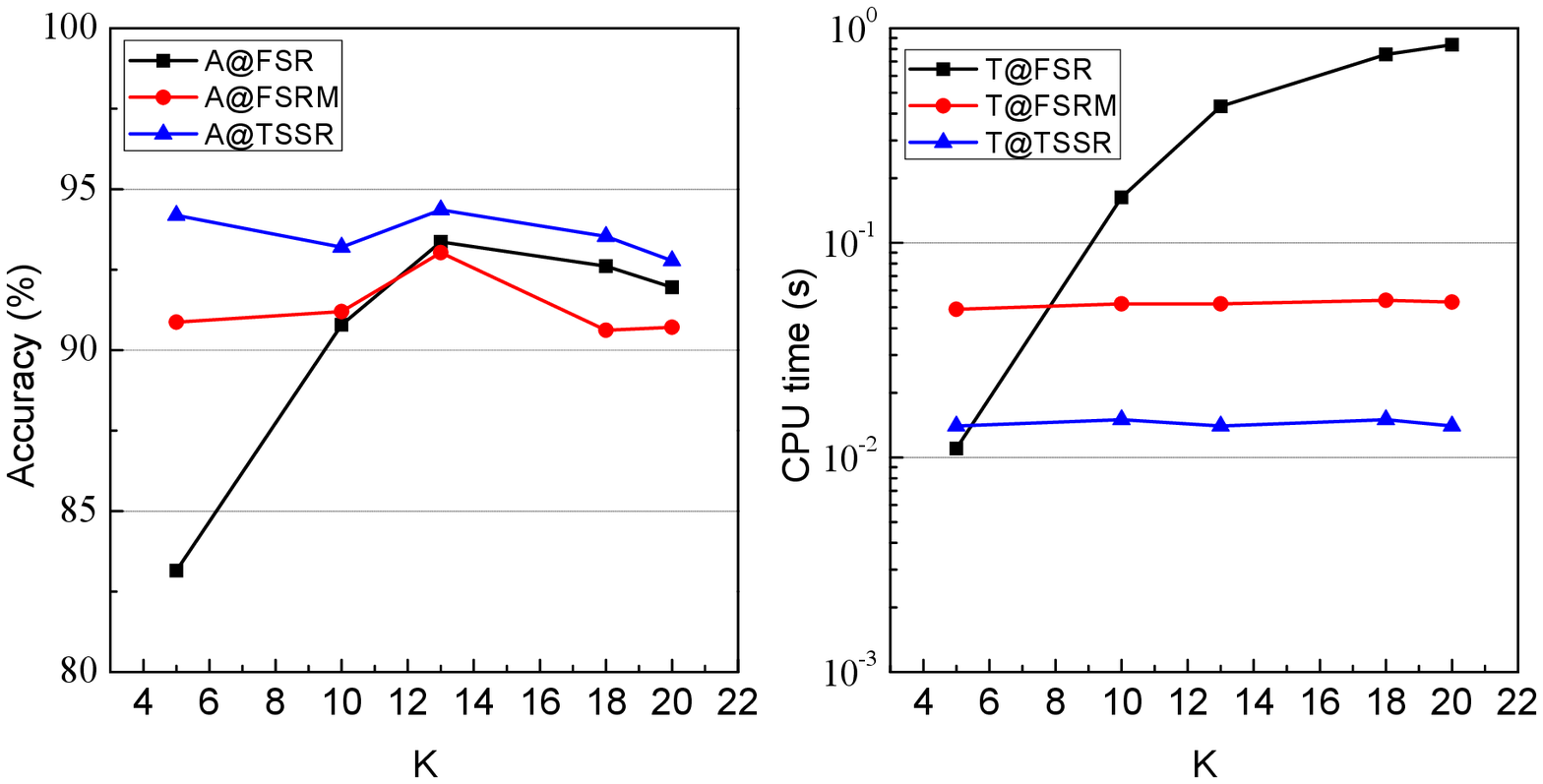}
\caption{The sparsity level K selection using FSR, FSRM and TSSR}
\label{fig_2}
\end{figure}

 We compare TSSR with FSRM and FSR \cite{JiaBinHuang2010FSR} to evaluate their performance since they are all algorithms with two-stage structure and using sparsity level $K$ to represent coefficients. For FSR, the KSVD algorithm \cite{Aharon05k-svd} within it is constrained
 by the dimension of training vectors, and trivial solution would occur if the dimension of the training vector is
 larger than the size of the dictionary. So we performed all the experiments on the extended YaleB database, using half of samples for training and the rest for testing as FSR did \cite{JiaBinHuang2010FSR}. The speed comparison between FSR and TSSR and the accuracy of all the algorithms under different $K$ can be seen in Table 4 and Figure \ref{fig_2}.
 We can find:

1. TSSR is more accurate than the other two methods. Interestingly for the methods themselves, there is a trade-off
between recognition accuracy and speed. Each method obtains its highest accuracy at the predetermined
value $K$, i.e., 13.2. This may be due to precondition that the normalization of the original data satisfies the
requirement of Gaussian distribution, and Eqn. (\ref{eqn_18}) does its work.

2. The TSSR is the fastest among all the methods in most cases. As $K$ increases, the run time ratio of FSR to TSSR becomes larger,
which is from 1 ($K=5$) to 61 ($K=20$). FSR needs a little less time than TSSR when $K=5$, which is much lower in accuracy than the others.

3. FSRM runs faster than FSR for most cases. This is because KSVD adopted in FSR needs much
time to generate coefficients for the second coding, while FSRM feeds the adopted $L_1$-Magic
with a much smaller input set in a two-stage process. Meanwhile, TSSR needs almost the same time to run for every $K$,
which is similar to FSRM. The reason is that TSSR and FSRM have similar two-stage structure
and both encode coefficients sparse enough which are generated in the first stage.

\subsection{Discussion}

Extensive experiments on four biometrics databases have revealed some significant points, from which we can find the following:

1. Comparing with several classical methods for solving linear inverse problems, experiments on PIE, AR and PolyU palmprint databases show that TSSR is an effective and efficient method.

2. TSSR almost uses the same time for different $K$ to run. This indicates the robustness of algorithm, which is not sensitive to the parameter $K$. This happens as long as we can obtain a small number of good discriminative features in the first stage to do the second sparse coding.

3. Since TSSR reaches the highest accuracy at about the predetermined value $K$, we can first use Eqn. (\ref{eqn_18}) to get the initial $K$ for a dataset, then adjust $K$ in experiments to acquire the desired results. This may give us freedom and avoid trial and error.

4. ALB and FISTA can obtain quite high accuracy for all the three databases, which can be used in the situation when highest accuracy is desired and we can afford the time.

5. TSSR structure can accommodate different sparse solvers as well as different measure matrices, if $\Psi$ is designed to obey or nearly obey Gaussian distribution with the training data, while $\Omega$ approximates an identity matrix.

6. Since the sparsity basis $\Omega$ is approaching identity matrix but not exactly, this may have effect on recognition rate. So the design of $\Omega$ is worth studying further.

\section{Conclusions and Future Work}
We have proposed a new method of Two-Stage Sparse Representation (TSSR) for solving linear inverse problem. TSSR makes the sparse approximation computationally tractable without sacrificing stable convergence.
Experimental results with image recognition indicate TSSR is more efficient with comparable accuracy than
several classic methods solving linear inverse problem.
As the proposed method provides a good way to exploit the special structure of biometric datasets and is helpful for achieving
rapid speed, it can be also applied to other recognition tasks.

\appendix

\section{Concentration of Measure}\label{apd:first}

Suppose we have a signal $y\in\mathbb{R}^m$, and a matrix $\Phi\in\mathbb{R}^{m\times N}$ in which the signal is presented.
We want to obtain a sparse set of coefficients (sparsity level K) to represent the signal. Assume that  an observed
sample belongs to one certain class and can be well represented using samples from the same class.
We say that the matrix $\Phi$ satisfies RIP of order $K$ with constant $\delta=\delta_K<1$ if
\begin{equation}
\label{eqn_19} \|x\|_0\leq K\Rightarrow(1-\delta_K)\|x\|_2^2\leq\|\Phi x\|_2^2\leq(1+\delta_K)\|x\|_2^2.
\end{equation}

RIP is a measure of closeness to an identity matrix for sparse vectors. Define that $\Phi_K$ selects $K$ columns
from $\Phi$, the RIP then suggests that every $\Phi_K$ should behave like an isometry - not changing the length of
the vector it multiplies. If $\|x\|_0$ is small enough, the norm $\|\Phi x\|_2^2$ can be constrained to a
small enough value, in which case a sparse representation can be stably determined \cite{candes2008restricted}.

We can generate random $m\times N$ matrices $\Phi$ by choosing the entries $\phi_{ij}$ as independent and identically distributed (i.i.d.) random variables.
For any $x\in\mathbb{R}^N$, the random variable $\|\Phi\|_2^2$ is strongly concentrated about its expected value, that is,
\begin{equation}
\label{eqn_20} Pr(|\|\Phi x\|_2^2-\|x\|_2^2|\geq\alpha\|x\|_2^2)\leq2e^{-cm\alpha^2},
\end{equation}
where the probability is taken over $\Phi$ and $c$ is a constant, for any $\alpha \in (0,1)$. This is called concentration of measure inequalities.

Any distribution satisfying a concentration inequality will provide both the canonical RIP and the universality with respect to a certain sparsifying basis $\Psi$. Here $\Psi$ is represented as a $N\times N$ identity matrix and $x\in\Sigma_K$, with which we can acquire signals $\Psi x$ that are sparse or compressible in practice.
By choosing $K$-dimensional subspaces spanned by sets of $K$ columns of $\Psi$, theorem 5.2 of \cite{baraniuk2008simple} establishes the RIP for $\Phi\Psi$ for each of the distributions.
See \cite{Davenport2011Sub-Gaussian} for more details on Sub-Gaussian random variables.

If the matrix $\Phi\Psi$ satisfies the RIP of order $2K$, then  $\Phi$ is a $\delta$-stable embedding of
$(\Psi(\sum_K),\Psi(\sum_K))$, where $\Psi(\sum_K)=\{\Psi x:x\in\sum_K\}$. We would require that $\Phi\Psi$ satisfies the RIP, and thus
bound the error  $\|\hat{x}-x\|_2$ introduced by the embedding. See the appendix B.

\section{Bi-Lipschitz Theory}\label{apd:second}

\textbf{Theorem 1}
 Suppose that $\Phi$ satisfies the RIP of order 2$K$ with isometry constant $\delta<\sqrt{2}-1$. Given measurements
 of the form $y=\Phi x+e$, where $\|e\|_2\leq\epsilon$, the solution $\hat{x}$ to
\begin{equation}
\label{eqn_21} \arg \min_{x'\in R^N}\|x'\|_1 \  \mathrm{s.t.} \ \|\Phi x'-y\|_2\leq\epsilon
\end{equation}
obeys
\begin{equation}
\label{eqn_22} \| \hat{x} - x\|_2\leq C_0\epsilon + C_1\frac{\|x - x_K\|_1}{\sqrt{K}},
\end{equation}
where
$C_0 =4 {\frac{\sqrt{1 + \delta}}{1 - (1 + \sqrt{2}) \delta}}, \ C_1 = 2{\frac{1 - (1 - \sqrt{2}) \delta}{1 - (1 + \sqrt{2}) \delta}}$,
$x_K$ is denoted as the vector $x$ with all but the $K$-largest entries set to zero.
We can restate the RIP in a more general form. Let $\delta\in(0,1)$ and $U,V\in\mathbb{R}^N$  be given, we say a mapping $\Phi$ is a $\delta$-stable embedding of $(U,V)$ if
\begin{equation}
\label{eqn_23} (1-\delta)\|u-v\|_2^2\leq\|\Phi u-\Phi v\|_2^2\leq(1+\delta)\|u-v\|_2^2,
\end{equation}
for all $u\in U$, $v\in V$. A mapping satisfying the property is commonly called \emph{bi-Lipschitz} \cite{baraniuk2008simple}.

\textbf{Lemma 1 \cite{davenport2010signal}:} Let $U$ and $V$ be sets of points in $\mathbb{R}^N$. Fix $\alpha,\beta\in(0,1)$. Let $\Phi$ be an $m\times N$ random matrix with \emph{i.i.d.} entries chosen from a distribution satisfying Eqn. (\ref{eqn_21}). If
\begin{equation}
\label{eqn_24} m\geq\frac{\ln(|U||V|)+\ln(2/\beta)}{c\alpha^2},
\end{equation}
then with probability exceeding $1-\beta$, $\Phi$ is a $\delta$-stable embedding of $(U,V)$. With Lemma 1, we can derive that $\Phi$ is a $\delta$-stable embedding of $(y_O,\{\Phi_O x\})$.

\end{document}